\documentclass[]{malab}

\usepackage{xspace}

\makeatletter
\DeclareRobustCommand\onedot{\futurelet\@let@token\@onedot}
\def\@onedot{\ifx\@let@token.\else.\null\fi\xspace}

\def\ie{\emph{i.e}\onedot}

\makeatother

\usepackage{multirow}
\usepackage{amsfonts}
\usepackage[normalem]{ulem}
\useunder{\uline}{\ul}{}

\newcommand{\tocite}[1]{\textcolor{red}{[TO CITE]}}

\usepackage{enumitem}

\usepackage[table]{xcolor}
\definecolor{tablegroupgray}{HTML}{EFEFEF}

\title{RUTA: Principled Visual Token Allocation via Rate-Utility Optimization}
\shorttitle{RUTA: Rate-Utility Visual Token Allocation}

\author[1]{Jian Zou}
\author[1]{Xiaoyu Xu}
\author[1]{Zhihua Wang}
\author[2]{Yilin Wang}
\author[2]{Balu Adsumilli}
\author[1]{Kede Ma}
\affiliation[1]{City University of Hong Kong}
\affiliation[2]{Google Inc.}

\abstract{
High-resolution images and long videos provide vision-language models with rich context for multimodal reasoning and fine-grained perception, but the resulting long visual token sequences make large language model-side computation and memory costly. Existing visual token reducers often operate at prescribed rates, while recent methods adapt token counts across inputs using method-specific learned thresholds or importance predictors. We introduce RUTA, a principled \textbf{R}ate-\textbf{U}tility \textbf{T}oken \textbf{A}llocation method that performs pre-LLM reduction by jointly learning which tokens to retain and how many to allocate to each image-query pair. RUTA constructs query-conditioned candidate tokens and predicts a retention probability for each candidate. During training, these probabilities parameterize independent Bernoulli gates, while their sum provides a differentiable training-time estimate of the token count for each pair. Retained tokens serve as anchors that aggregate information from non-retained tokens according to semantic affinity and spatial proximity. RUTA is optimized with a penalized rate-utility objective that balances downstream task loss against expected token usage. Averaged across five benchmarks and measured relative to each backbone's full-token baseline, RUTA uses only $2.0\%$ and $4.2\%$ of visual tokens while preserving $88.2\%$ and $94.4\%$ of task performance on LLaVA-NeXT-7B and Qwen3-VL-8B, respectively.
}

\date{\today}
\authoremails{
\email{jian.zou@my.cityu.edu.hk},
\email{\{xiaoyxu, zhihua.wang\}@cityu.edu.hk},
\email{\{yilin, badsumilli\}@google.com}
}
\correspondence{\email{kede.ma@cityu.edu.hk}}
\projectpage{\url{https://github.com/Multimedia-Analytics-Laboratory/RUTA}}

\begin{document}

\maketitle


\section{Introduction}

Visual intelligence depends on the fidelity and scope of the evidence available for inference~\citep{marr1982vision}. Accordingly, vision-language models (VLMs) increasingly process high-resolution images and long videos to resolve fine-grained visual details and provide broader spatial and temporal context for multimodal reasoning~\citep{guo2024llava,li2024monkey,wang2024qwen2}. These gains, however, require a corresponding expansion of the visual representation: richer inputs produce longer token sequences that must be processed by the large language model (LLM). In multi-image and long-video settings, the resulting token sequence growth substantially increases LLM-side computation and memory costs~\citep{wu2024longvideobench,zhang2025mme}, making visual token efficiency a central constraint on scalable multimodal reasoning.

Visual token reduction addresses this cost by pruning or merging tokens before or during LLM-side processing. The central premise is that useful visual evidence is unevenly distributed across the spatial and temporal extent of visual inputs: answering a query (\ie, a question or instruction about the visual input) may require evidence from only a small region, while large portions of the input remain irrelevant or redundant~\citep{chen2024efficient,chen2024image,yang2025visionzip}. Existing reducers exploit this structure using query-conditioned relevance~\citep{zhang2025sparsevlm}, redundancy-aware pruning or merging~\citep{Shang_2025_ICCV,yang2025visionzip}, diversity-preserving selection~\citep{alvar2025divprune}, or progressive layer-wise reduction~\citep{xing2025conical}.

\begin{figure}[!t]
	\centering
	\includegraphics[width=.85\textwidth]{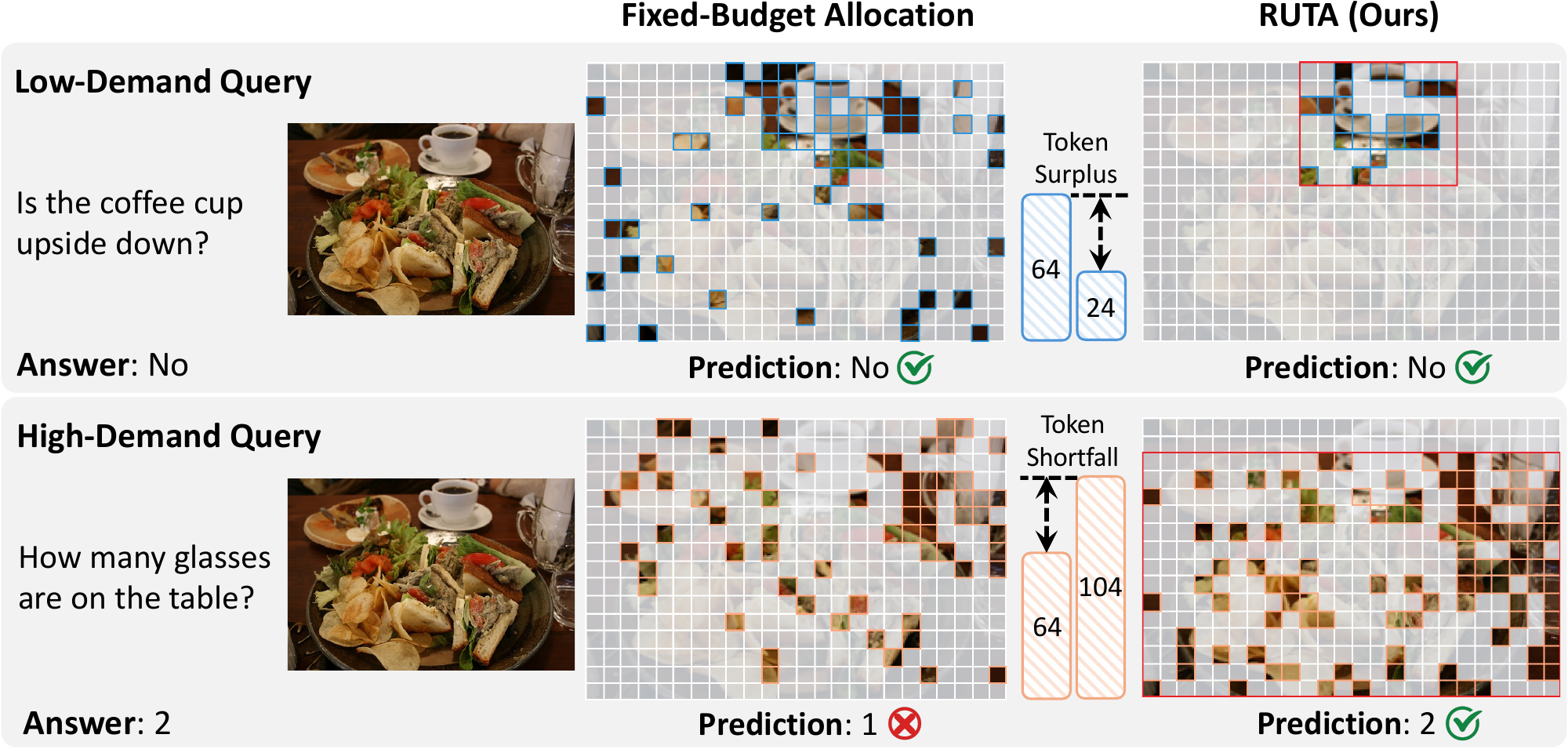}
	\caption{
    Fixed-budget versus adaptive visual token allocation. A fixed budget retains $64$ tokens for each query, leaving surplus capacity for the low-demand query but insufficient evidence for the high-demand query, which leads to an incorrect answer. At the same average rate, RUTA retains $24$ and $104$ tokens, respectively, adapting token usage to each query's evidence demand and answering both correctly. Red boxes mark the query-conditioned regions of interest.
    }
	\label{fig:teaser}
\end{figure}

Despite these advances, many reducers continue to rely on externally specified token counts, retention ratios, or layer-wise schedules~\citep{yang2025visionzip,alvar2025divprune,xing2025conical}. These controls make computation predictable, but a uniform per-sample budget may overallocate tokens to simple queries while underallocating them to queries requiring broader visual evidence, as illustrated in Fig.~\ref{fig:teaser}. Recent methods have therefore begun to vary token usage across inputs or LLM layers using learned thresholds or importance predictors~\citep{ye2025atp,zeng2025glimpse}. These studies demonstrate the value of adaptive allocation, but their rate control mechanisms remain tied to method-specific objectives and constraints, motivating a unified formulation that explicitly balances token usage against downstream performance.

Drawing inspiration from the rate-distortion theory~\citep{shannon1959coding}, we formulate visual token reduction as a rate-utility optimization problem. Rate is measured by the expected number of visual tokens passed to the LLM, whereas utility is represented by downstream task performance. Rather than imposing the same token budget on every image-query pair, we apply a shared penalty to expected token usage. This objective allows the reduction policy to assign different token counts across inputs, while the rate coefficient controls the average rate-utility operating point.

We instantiate this formulation with RUTA, a principled \textbf{R}ate-\textbf{U}tility \textbf{T}oken \textbf{A}llocation method. RUTA performs token reduction before LLM processing, allowing token usage to vary with the visual evidence required by each input query.
More specifically, RUTA first constructs query-conditioned visual token candidates and assigns each candidate a retention probability, from which discrete token selections are sampled independently during training. The sum of these probabilities gives the expected number of retained tokens for each sample. Retained tokens are treated as anchors to absorb information from non-retained tokens (\ie, non-anchors) according to semantic affinity and spatial proximity. The trainable candidate construction, allocation, and aggregation components in RUTA are jointly optimized via a rate-utility objective that combines downstream task loss with a penalty on expected token usage.

We evaluate RUTA on five different benchmarks using LLaVA-NeXT-7B~\citep{liu2024llavanext} and Qwen3-VL-8B~\citep{bai2025qwen3}. At matched low-token rates, RUTA achieves the highest average task performance among the evaluated reducers on both backbones. In particular, RUTA uses only $2.0\%$ and $4.2\%$ of visual tokens while preserving $88.2\%$ and $94.4\%$ of task performance relative to each backbone's full-token baseline on LLaVA-NeXT-7B and Qwen3-VL-8B, respectively.

In summary, our key contributions are threefold.
\begin{itemize}
    \item [1.] We formulate visual token reduction as rate-utility optimization, explicitly balancing downstream task performance against expected token usage.
    \item [2.] We introduce RUTA, which integrates query-conditioned candidate construction, Bernoulli retention, and anchor-based token aggregation, within a jointly trained pre-LLM reduction framework.
    \item [3.] We demonstrate consistent gains by RUTA at matched low-token rates and favorable rate-utility trade-offs under more severe token reduction.
\end{itemize}

\section{Related Work}

In this section, we position RUTA along two complementary lines of visual token reduction: training-free methods that operate without additional model optimization and training-based methods that learn compact token representations or reduction policies.

\subsection{Training-Free Visual Token Reduction}

Training-free methods can be organized by whether reduction occurs before or within LLMs. Pre-LLM methods compact the sequence within the vision encoder or at its output, including adaptive token sampling~\citep{fayyaz2022adaptive} and similarity-based token merging~\citep{bolya2023token}. In VLMs, importance-based methods retain visually salient or informative tokens~\citep{Shang_2025_ICCV,yang2025visionzip,zhang2025beyond}; diversity-aware methods seek broader visual coverage~\citep{alvar2025divprune}; and PruneSID~\citep{fang2026prune} dynamically adjusts the reduction ratio according to estimated image complexity. By compacting the sequence before LLM ingestion, these methods avoid LLM-side computation on removed or merged tokens and remain broadly compatible with existing VLMs. For most of these methods, however, the operating rate is specified through an external token count or retention ratio.

In-LLM methods instead defer reduction until multimodal processing has begun. FastV~\citep{chen2024image} identifies substantial redundancy after shallow LLM layers; SparseVLM~\citep{zhang2025sparsevlm} performs text-guided, rank-based layer-adaptive sparsification; ZipVL~\citep{He_2025_ICCV} assigns layer-wise sparsity; PyramidDrop~\citep{xing2025conical} progressively drops tokens across layers; and VScan~\citep{zhang2026vscan} combines global and local visual scanning. Thus, visual token usage can vary across LLM layers even in training-free methods, although the resulting rates are determined by heuristic rules or externally selected operating parameters rather than optimized jointly with downstream task utility.

\subsection{Training-Based Visual Token Reduction}

Training-based methods use task supervision to learn content-dependent visual computation. In vision Transformers, early methods either aggregate informative content into a small token set or progressively bypass less useful tokens~\citep{ryoo2021tokenlearner,rao2021dynamicvit,yin2022vit}. These methods establish adaptive visual processing, but were designed primarily for recognition pipelines rather than VLMs.

In VLMs, learned efficiency mechanisms operate at several points between the input image and the LLM. One family uses latent queries to map variable-length visual features into a fixed-capacity interface~\citep{jaegle2021perceiver,jaegle2022perceiver,alayrac2022flamingo,li2023blip,dai2023instructblip}. Other methods improve how spatial details from high-resolution inputs are organized and aggregated~\citep{cha2024honeybee,tong2024cambrian,guo2024llava,cai2025matryoshka,li2025tokenpacker}. Efficient vision encoders and compact multimodal interfaces further lower visual encoding cost or LLM input length~\citep{vasu2025fastvlm,zhang2025llava}. Together, these designs improve efficiency, but the operating rate is typically built into the interface or selected externally rather than learned jointly with task utility.

Modular trainable reducers preserve most of the base VLM. TwigVLM~\citep{shao2025growing} adds a shallow branch to guide visual token pruning and accelerate text generation, while CROP~\citep{guo2025crop} localizes query-relevant context to guide either pre-LLM reduction or in-LLM pruning. Although supervision makes their selection decisions task-aware, their operating rates generally remain governed by prescribed token counts or retention settings.

More recent trainable reducers explicitly adapt visual token counts across inputs. ATP-LLaVA~\citep{ye2025atp} predicts sample- and layer-specific pruning thresholds inside the LLM and optimizes them using task, computation, and target-count losses. GlimpsePrune~\citep{zeng2025glimpse} applies a learned importance predictor after several LLM layers to produce a dynamic one-shot pruning mask, optionally subject to a maximum retention ratio. Together, these methods establish adaptive token allocation as an important design dimension, while relying on method-specific objectives and constraints to determine the operating rate. RUTA builds on this line of work by explicitly formulating visual token reduction as rate-utility optimization, and differs by directly optimizing a differentiable expected pre-LLM token rate and coupling this allocation with query-conditioned candidate construction and anchor-based aggregation.

\section{Proposed Method: RUTA}

\begin{figure}[!t]
	\centering
	\includegraphics[width=.9\textwidth]{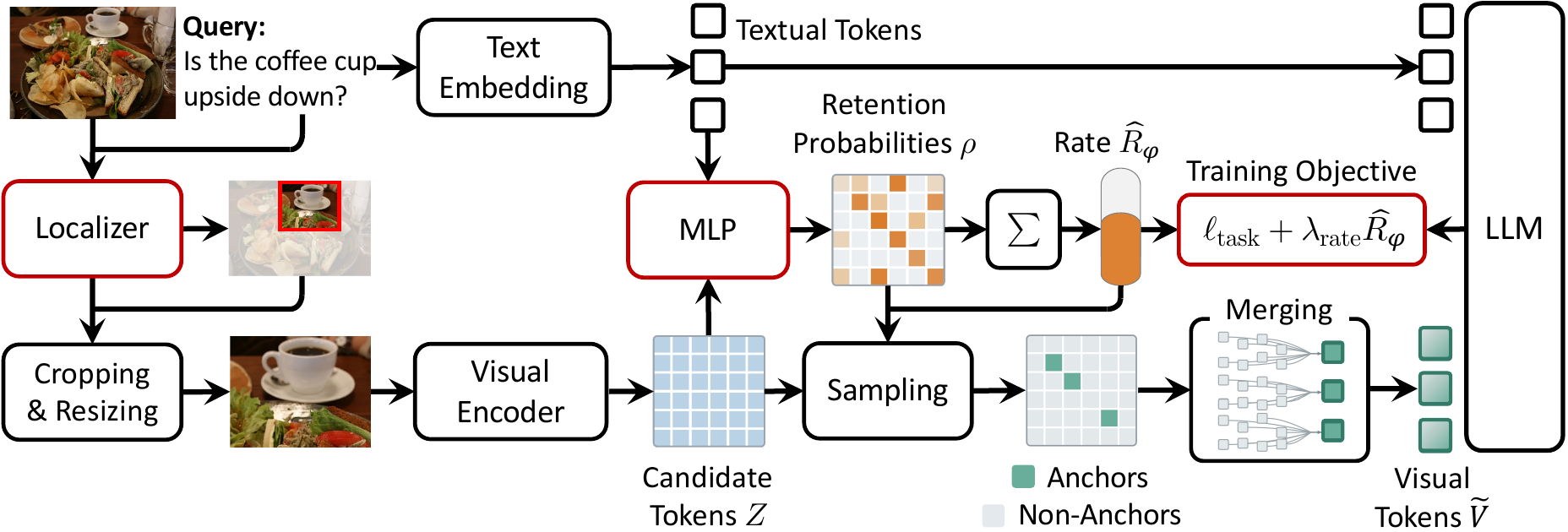}
	\caption{System diagram of RUTA. Given an image and a query, a query-conditioned localizer identifies a region of interest, which is cropped, resized, and encoded into candidate visual tokens \(Z\). A two-layer MLP predicts a retention probability \(\rho_i\) for each candidate, and their sum gives the expected token rate \(\widehat{R}_{\varphi}\). During training, independent Bernoulli sampling partitions the candidates into anchors and non-anchors. The non-anchor information is then merged into the anchors to produce the compact visual token sequence \(\widetilde{V}\). RUTA is trained with a task loss and a rate penalty weighted by \(\lambda_{\mathrm{rate}}\), balancing prediction utility against the number of visual tokens passed to the LLM.
    }
	\label{fig:overview}
\end{figure}

In this section, we introduce RUTA, which uses rate-utility optimization to jointly determine which visual tokens to retain and how many to allocate to each image-query pair. We measure the operating rate by the average number of visual tokens passed to the LLM across image-query pairs and utility by downstream task performance. The system diagram of RUTA is shown in Fig.~\ref{fig:overview}.

\subsection{Problem Formulation}
\label{sec:problem_formulation}

Let $(x,q,y)$ denote an input image, a textual query, and its target output. A standard VLM maps $x$ through a vision encoder \(E_\mathrm{vis}\) to a dense sequence of $N$ visual tokens:
\(V=E_\mathrm{vis}(x)=[v_i]_{i=1}^{N}\), where \(v_i\in\mathbb{R}^{D}\).
The LLM conditions on both $q$ and $V$ to predict $y$. In the pre-LLM setting considered here, let $K(x,q)\leq N$ denote the number of visual tokens passed to the LLM for an image-query pair. For a pre-trained VLM, all visual tokens entering the LLM have the same embedding dimension \(D\). Although different layers may use different bit depths under a static mixed-precision inference configuration, the layerwise precision assignment is fixed across inputs~\citep{dong2019hawq,shen2020qbert,yao2022zeroquant,frantar2023optq}. \textit{We therefore use token count as a model-relative measure of rate and a proxy for LLM-side computation and memory}.

At the LLM interface, a pre-LLM reduction policy $f_\theta$, parameterized by vector $\theta$, replaces the dense encoder output $V$ with an input-dependent compact sequence:
\begin{equation}
\widetilde{V}
=
f_\theta(x,q)
=
[\tilde{v}_i]_{i=1}^{K(x,q)}.
\label{eq:compact_visual_tokens}
\end{equation}
Here, $f_\theta$ denotes the complete visual processing path before the LLM and may include or reuse $E_{\mathrm{vis}}$. The LLM receives its output $\widetilde{V}$ in place of the baseline dense sequence $V$. Because the token count \(K(x,q)\) depends on the image-query pair, the policy can allocate more tokens when broader visual evidence is required and fewer tokens when the relevant evidence is concentrated.

An in-LLM reducer instead operates within the LLM and may process different numbers of visual tokens across layers. If \(K_l(x,q)\) denotes the number of visual tokens processed at layer \(l\) in an LLM with \(L\) layers, its rate can be summarized by the layer-wise average $\frac{1}{L}\sum_{l=1}^{L}K_l(x,q)$.

Following the rate-distortion theory~\citep{shannon1959coding}, we use a differentiable token rate estimate $\widehat{R}_{\theta}(x,q)$ and solve the following rate-utility optimization problem:
\begin{equation}
\min_{\theta}
\mathbb{E}_{(x,q,y)}
\left[
\ell_{\mathrm{task}}
\bigl(\mathrm{LLM}(q,f_\theta(x,q)),y\bigr)
+
\lambda_{\mathrm{rate}}
\widehat{R}_{\theta}(x,q)
\right],
\label{eq:generic_rate_utility_objective}
\end{equation}
where $\lambda_{\mathrm{rate}}$ controls the trade-off between the two terms. 

RUTA further decomposes the reduction policy $f_\theta$ into a query-conditioned candidate constructor $g_{\phi}$ and a rate-controlled compressor $h_{\varphi}$:
\begin{equation}
Z=g_{\phi}(x,q), 
\qquad
\widetilde{V}=h_{\varphi}(Z,q),
\label{eq:ruvtr_functions}
\end{equation}
where $\theta=(\phi,\varphi)$ collects the trainable parameters of the candidate constructor and compressor. Consequently, $f_\theta(x,q)=h_{\varphi}(g_{\phi}(x,q),q)$ and $\widehat{R}_{\theta}(x,q)=\widehat{R}_{\varphi}(g_{\phi}(x,q),q)$. The constructor $g_{\phi}$ concentrates the candidate sequence on query-relevant visual content, while the compressor $h_{\varphi}$ assigns each candidate a retention probability, which guides anchor selection. The remaining candidates are then aggregated into the selected anchors to form the compact sequence passed to the LLM. The sum of the same probabilities also defines the sample-specific expected rate.

\subsection{Query-Conditioned Candidate Token Construction}
\label{sec:candidate_construction}

The candidate constructor $g_{\phi}$ first identifies a query-relevant image region, resizes it to the vision encoder’s standard input resolution, and encodes it into candidate tokens. This concentrates the candidate tokens on relevant visual evidence before token reduction.
Specifically, given the query $q$, RUTA extracts $R(q)$ phrases for grounding relevant image regions:
\begin{equation}
\{u_j\}_{j=1}^{R(q)} = \mathrm{parser}(q).
\label{eq:query_phrase_extraction}
\end{equation}
The $\mathrm{parser}(\cdot)$ removes question templates and other non-visual words, retaining noun phrases and attributes that describe image content. A phrase-conditioned localizer $d_\phi$ then returns a bounding box $b_j$ and confidence score $c_j$ for each phrase $u_j$:
\begin{equation}
\{(b_j,c_j)\}_{j=1}^{R(q)}
=
d_{\phi}\left(x,\{u_j\}_{j=1}^{R(q)}\right).
\label{eq:phrase_conditioned_localization}
\end{equation}
The localizer $d_{\phi}$ may be instantiated by any open-vocabulary grounding model, and we adopt Grounding DINO~\citep{liu2024grounding} by default. Using confidence threshold $\delta$, RUTA forms a set of candidate boxes and, if this set is nonempty, takes their union as the query-conditioned region of interest:
\begin{equation}
\mathcal{B}_{\delta}=\{b_j \mid c_j \ge \delta\},
\qquad b_{\mathrm{roi}}=\mathrm{Union}(\mathcal{B}_{\delta}).
\label{eq:focus_box}
\end{equation}
Here, \(\operatorname{Union}(\cdot)\) returns the smallest axis-aligned bounding box enclosing all boxes in \(\mathcal B_\delta\). We use a single enclosing region rather than processing multiple regions separately because it preserves potentially useful \textit{intervening} evidence and the spatial relationships among the localized regions, while preventing the number of encoder views---and hence candidate tokens---from scaling with the number of grounded regions. When $\mathcal{B}_{\delta}=\emptyset$, RUTA skips cropping and passes the full image to the vision encoder. RUTA resizes the selected content to the base VLM input resolution and encodes it as the candidate sequence $Z$:
\begin{equation}
Z
=
g_{\phi}(x,q)
=
E_{\mathrm{vis}}
\bigl(
\mathrm{Resize}(\mathrm{Crop}(x,b_\mathrm{roi}))
\bigr)
=
[z_i]_{i=1}^{N_c},
\qquad
z_i\in\mathbb{R}^{D}.
\label{eq:candidate_visual_tokens}
\end{equation}
Because the crop is resized to the base resolution, candidate construction does not necessarily shorten the visual sequence: $N_c$ remains comparable to the original dense token count $N$.\footnote{For any-resolution or tiled inputs, $N_c$ may differ from $N$, depending on the number of encoded views.} Instead, it reallocates the encoder's spatial resolution to query-relevant content, while the subsequent compressor $h_{\varphi}$ performs the actual token reduction.

\subsection{Rate-Controlled Compression}
\label{sec:rate_modeling}

Given $Z=[z_i]_{i=1}^{N_c}$, the compressor $h_{\varphi}$ performs three steps: it predicts a retention probability for each candidate, samples independent binary gates, and merges information from non-retained tokens into the retained anchors. The probabilities jointly determine which tokens are preferred and how many are expected to remain. Within the compressor \(h_\varphi\), a two-layer query-conditioned multilayer perceptron (MLP) $s_{\varphi}$ induces each candidate a retention probability:
\begin{equation}
\rho_i
=
\mathrm{sigmoid}\left(\frac{s_{\varphi}(z_i,q)}{\tau}\right), \qquad i \in \{1, \ldots, N_c\},
\label{eq:retention_probability}
\end{equation}
where \(\tau>0\) is a temperature controlling the sharpness of these probabilities.  During training, RUTA samples an independent Bernoulli gate for every candidate and partitions the candidates into an anchor index set $\mathcal{A}$ and a non-anchor index set $\bar{\mathcal{A}}$:
\begin{equation}
\xi_i \sim \mathrm{Bernoulli}(\rho_i),
\qquad
\mathcal{A}=\{i\mid \xi_i=1\},
\qquad
\bar{\mathcal{A}}=\{i\mid \xi_i=0\}.
\label{eq:bernoulli_sampling}
\end{equation}
As $|\mathcal{A}|=\sum_i\xi_i$, the expected number of anchors under independent Bernoulli sampling is
\begin{equation}
\mathbb{E}[|\mathcal{A}|]
=
\sum_{i=1}^{N_c}\rho_i
=
\widehat{R}_{\varphi}(Z,q).
\label{eq:expected_visual_rate}
\end{equation}
This differentiable quantity serves as the training-time rate estimate, avoiding prescribed token budgets. If a Bernoulli draw selects  no anchor, yielding $\mathcal{A}=\emptyset$, RUTA sets \(\xi_{i^\star}=1\), where \(i^\star=\operatorname*{arg\,max}_i\rho_i\), thereby ensuring a nonempty anchor set.

Because gradients cannot be usefully propagated through the discrete Bernoulli samples, RUTA uses a straight-through estimator~\citep{bengio2013estimating, van2017neural}:
\begin{equation}
\tilde{\xi}_i
=
\rho_i+\operatorname{sg}(\xi_i-\rho_i),
\qquad
i\in\{1,\ldots,N_c\},
\label{eq:st_anchor_gate}
\end{equation}
where \(\operatorname{sg}(\cdot)\) denotes the stop-gradient operator. Thus, \(\tilde{\xi}_i\) equals the discrete sample \(\xi_i\) in the forward pass, while its gradient with respect to \(\rho_i\) is set to one in the backward pass.

At inference, the learned probabilities can be converted into token selections either stochastically, through Bernoulli sampling, or deterministically, through fixed thresholding or adaptive top-$K$. We use adaptive top-$K$ by default and set
\begin{equation}
K_{\varphi}(Z,q)
=
\operatorname{clip}\!\left(
\operatorname{round}\!\left(\widehat{R}_{\varphi}(Z,q)\right),
K_{\min},N_c
\right),
\label{eq:adaptive_topk_inference}
\end{equation}
where \(K_{\min}\in\{1,\ldots,N_c\}\) denotes the minimum allowable number of retained tokens. We construct the anchor index set \(\mathcal A\) from the indices of the \(K_\varphi(Z,q)\) candidates with the highest retention probabilities, and assign all remaining indices to the non-anchor index set \(\bar{\mathcal A}\). This policy converts the learned expected rate into a sample-specific, deterministic token count.

Discarding the non-anchors would remove potentially useful context. RUTA instead distributes information from each non-anchor across the retained anchors according to semantic affinity and spatial proximity. For each anchor \(z_i\) indexed by \(i\in\mathcal A\) and each non-anchor \(z_j\) indexed by \(j\in\bar{\mathcal A}\), we first compute an unnormalized compatibility score:
\begin{equation}
e_{i,j}
=
\frac{
\left\langle
W_{\mathrm{query}} z_i,
W_{\mathrm{key}} z_{j}
\right\rangle
}{\sigma_\mathrm{sa}}
-
\lambda_{\mathrm{sp}}
\frac{\|\mathrm{pos}_i-\mathrm{pos}_j\|_2^2}{2\sigma_{\mathrm{sp}}^2}.
\label{eq:anchor_merge_assignment_1}
\end{equation} 
The first term in $e_{i,j}$ measures learned semantic content affinity by mapping tokens through $W_\mathrm{query}$ and $W_\mathrm{key}$ into a shared affinity space. The scale $\sigma_{\mathrm{sa}}>0$ controls the sharpness of this affinity. The second term favors nearby tokens, where $\mathrm{pos}_i\in[0,1]^2$ is the normalized spatial center of token $z_i$ and $\sigma_{\mathrm{sp}}>0$ is the spatial bandwidth. $\lambda_{\mathrm{sp}}\geq0$ controls the strength of the spatial bias.
For each \(j\in\bar{\mathcal A}\), we then normalize the scores \(\{e_{i,j}\}_{i\in\mathcal A}\) over the anchor indices to obtain 
\begin{equation}
w_{i,j}
=
\frac{\exp(e_{i,j})}
{\sum_{i'\in\mathcal{A}}\exp(e_{i',j})},
\label{eq:anchor_merge_assignment_2}
\end{equation} 
such that $\sum_{i\in\mathcal{A}}w_{i,j}=1$ for each non-anchor. RUTA then adds non-anchor information to each anchor:
\begin{equation}
\Delta z_i
=
\frac{
\sum_{j\in \bar{\mathcal{A}}} w_{i,j} W_\mathrm{value} z_j
}{
\max\left(1,\sum_{j\in \bar{\mathcal{A}}}w_{i,j}\right)
}, \qquad \tilde{z}_i
=
\tilde{\xi}_i\cdot\operatorname{LN}\left(z_i+\beta\Delta z_i\right),
\quad i\in \mathcal{A} .
\label{eq:anchor_residual_merge}
\end{equation} 
Here, $W_\mathrm{value}$ is a learned value projection with output dimension $D$, $\operatorname{LN}(\cdot)$ denotes layer normalization over the embedding dimension, and $\beta$ scales the residual update. The denominator in $\Delta z_i$ averages contributions when the total assignment weight exceeds one, while preserving the attenuation of weak assignments when the total weight is below one.

The learned query, key, and value projections follow the standard attention design and enable task-adaptive aggregation. Although our anchor selection does not explicitly enforce diversity, the subsequent aggregation mitigates potential coverage loss by routing information from every non-anchor. This allows complementary evidence beyond the selected anchors to remain represented in the compact sequence. The resulting $\widetilde{V}=[\tilde{z}_i]_{i\in \mathcal{A}}$ contains $K(x,q)=|\mathcal{A}|$ merged anchors and replaces $Z$ at the LLM input.

\subsection{Joint Rate-Utility Optimization}
\label{sec:joint_optimization}

RUTA jointly optimizes candidate construction and compression using task supervision and a differentiable rate penalty. Let
$\mathcal{D}=\{(x^{(i)},q^{(i)},y^{(i)})\}_{i=1}^{|\mathcal{D}|}$ denote a mini-batch of image-query-target triples. For each \((x,q,y)\in\mathcal D\), Eq.~(\ref{eq:ruvtr_functions}) produces \(Z\) and Bernoulli sampling produces the corresponding compact sequence \(\widetilde V\). Using one independently sampled gate vector per image-query pair in each
forward pass, we optimize the following stochastic mini-batch objective:
\begin{equation}
\ell(\mathcal{D};\phi,\varphi)
=
\frac{1}{|\mathcal{D}|}
\sum_{(x,q,y)\in\mathcal{D}}
\left[
\ell_{\mathrm{task}}
\left(
\mathrm{LLM}(q,\widetilde{V}),y
\right)
+
\lambda_{\mathrm{rate}}
\widehat{R}_{\varphi}(Z,q)
\right].
\label{eq:joint_rate_utility_loss}
\end{equation}
The task term teaches the model to retain and aggregate evidence needed for prediction, while the rate term penalizes the expected number of tokens passed to the LLM. The multiplier $\lambda_{\mathrm{rate}}$ controls the rate-utility operating point: larger values favor compact representations, whereas smaller values prioritize task utility. Unlike a fixed per-sample budget, this mini-batch-averaged objective permits different inputs to receive different token counts under the same global rate pressure.

\section{Experiments}

In this section, we evaluate whether RUTA improves task utility at a given visual token rate and whether its advantage persists across VLM backbones and operating points. Rate is measured as the number of visual tokens processed by the LLM, averaged over samples and layers. For a pre-LLM reducer, this reduces to the average visual input length; for an in-LLM reducer, it follows the layer-wise definition in Sec.~\ref{sec:problem_formulation}. Utility is measured by the standard accuracy metric for each benchmark.

\begin{table*}[!t]
\centering
\caption{
Matched low-rate results across five benchmarks.
Tok denotes the number of visual tokens averaged over samples and LLM layers, whereas Acc denotes benchmark-specific accuracy metric. Reference denotes performance without visual token reduction, and Mean Acc is the unweighted average across the five benchmarks. Boldface marks the highest accuracy among reduced methods within each backbone block.
}
\label{tab:main_results}
\resizebox{\textwidth}{!}{
\scriptsize
\begin{tabular}{l|cccccccccc|c}
\toprule
\multirow{2}{*}[-0.55ex]{{Method}} &
\multicolumn{2}{c}{{VQA v2}} &
\multicolumn{2}{c}{{GQA}} &
\multicolumn{2}{c}{{TextVQA}} &
\multicolumn{2}{c}{{A-OKVQA}} &
\multicolumn{2}{c|}{{VizWiz}} &
{Mean} \\
\cmidrule(lr){2-3}\cmidrule(lr){4-5}\cmidrule(lr){6-7}\cmidrule(lr){8-9}\cmidrule(lr){10-11}
& Tok & Acc & Tok & Acc & Tok & Acc & Tok & Acc & Tok & Acc & Acc \\
\midrule
\rowcolor{tablegroupgray}\multicolumn{12}{c}{\rule[-1.ex]{0pt}{3.2ex}{LLaVA-NeXT-7B}}\\
{\rule[-.5ex]{0pt}{3.ex}Reference} & 2,233 & 81.1 & 2,012 & 73.2 & 2,310 & 66.1 & 2,234 & 83.2 & 2,336 & 61.6 & 73.04 \\
\cmidrule(r){1-12}
FastV & 42.0 & 45.3 & 32.0 & 40.4 & 50.0 & 41.6 & 48.0 & 55.5 & 48.0 & 53.7 & 47.30 \\
SparseVLM & 42.1 & 68.3 & 32.0 & 58.0 & 50.9 & \textbf{51.0} & 48.3 & 75.5 & 48.3 & 57.1 & 61.98 \\
VisionZip & 42.1 & 48.7 & 31.0 & 41.7 & 49.8 & 45.9 & 48.9 & 61.5 & 49.7 & 53.9 & 50.34 \\
DivPrune & 41.9 & 67.5 & 32.0 & 53.2 & 50.0 & 49.6 & 48.0 & 75.7 & 47.7 & 58.4 & 60.88 \\
PruneSID & 41.6 & 55.7 & 33.3 & 48.0 & 49.6 & 30.6 & 48.8 & 67.6 & 49.7 & 52.7 & 50.92 \\
FastVLM & 49.0 & 73.0 & 36.0 & \textbf{64.2} & 49.0 & 46.7 & 49.0 & 54.6 & 49.0 & 58.7 & 59.44 \\
TwigVLM & 42.0 & 35.1 & 32.3 & 27.5 & 50.3 & 22.6 & 48.0 & 53.0 & 48.0 & 5.0 & 28.64 \\
\cmidrule(r){1-12}
RUTA (Ours) & 42.1 & \textbf{73.3} & 32.1 & 60.4 & 50.4 & 49.7 & 48.1 & \textbf{75.9} & 47.1 & \textbf{62.8} & \textbf{64.42} \\
\midrule
\rowcolor{tablegroupgray}\multicolumn{12}{c}{\rule[-1.ex]{0pt}{3.2ex}{Qwen3-VL-8B}}\\
{\rule[-.5ex]{0pt}{3.ex}Reference} & 1,341 & 83.8 & 1,343 & 72.3 & 1,337 & 82.3 & 1,341 & 89.0 & 1,307 & 68.7 & 79.22 \\
\cmidrule(r){1-12}
FastV & 30.0 & 43.2 & 70.0 & 51.5 & 70.0 & 52.2 & 70.0 & 65.8 & 40.0 & 56.5 & 53.84 \\
SparseVLM & 30.2 & 42.1 & 70.7 & 36.9 & 70.7 & 44.9 & 70.7 & 48.7 & 41.3 & 53.7 & 45.26 \\
VisionZip & 30.0 & 55.1 & 70.0 & 60.4 & 70.0 & 54.9 & 70.0 & 76.2 & 40.0 & 62.7 & 61.86 \\
DivPrune & 28.0 & 67.1 & 70.0 & 55.9 & 70.0 & 62.1 & 70.0 & \textbf{82.5} & 40.0 & 65.3 & 66.58 \\
PruneSID & 27.4 & 65.1 & 71.1 & 54.6 & 69.6 & 38.2 & 70.8 & 80.5 & 40.4 & 63.9 & 60.46 \\
FastVLM & 36.0 & 69.6 & 81.0 & 60.5 & 81.0 & 48.2 & 81.0 & 54.0 & 36.0 & 58.6 & 58.18 \\
TwigVLM & 30.0 & 71.2 & 70.0 & 62.8 & 70.0 & 62.6 & 70.0 & 68.9 & 40.0 & 59.4 & 64.98 \\
\cmidrule(r){1-12}
{RUTA} (Ours) & 28.8 & \textbf{76.6} & 69.4 & \textbf{71.9} & 70.4 & \textbf{73.1} & 70.1 & 81.8 & 39.7 & \textbf{70.5} & \textbf{74.78} \\
\bottomrule
\end{tabular}
}
\end{table*}

\subsection{Experimental Setups}
\paragraph{Datasets}
We use five benchmarks spanning complementary evidence demands: VQAv2~\citep{goyal2017making} for general visual question answering (VQA), GQA~\citep{hudson2019gqa} for compositional reasoning, TextVQA~\citep{singh2019towards} for scene-text understanding, A-OKVQA~\citep{schwenk2022okvqa} for knowledge-intensive reasoning, and VizWiz~\citep{gurari2018vizwiz} for questions collected from blind users in real-world settings. Following standard protocols, VQAv2, TextVQA, and VizWiz use soft accuracy, which assigns partial credit according to agreement among multiple plausible human answers. GQA uses exact-match accuracy, which awards credit only when the predicted answer matches the reference answer after standard answer normalization. A-OKVQA uses multiple-choice accuracy, measured as the percentage of questions for which the correct option is selected.

\paragraph{Implementation details}
We evaluate RUTA with LLaVA-NeXT-7B~\citep{liu2024llavanext} and Qwen3-VL-8B~\citep{bai2025qwen3}. The vision encoder, text embedding, and LLM remain frozen. Trainable components are limited to the grounding heads of the candidate localizer, the retention MLP, the anchor-merging projections in $h_{\varphi}$, and the VLM's lightweight vision-to-language projector. We set the confidence threshold \(\delta\) for phrase-conditioned boxes in Eq.~(\ref{eq:focus_box}) to \(0.5\), the retention temperature \(\tau\) in Eq.~(\ref{eq:retention_probability}) to \(0.7\), and the minimum token count $K_\mathrm{min}$ in Eq.~(\ref{eq:adaptive_topk_inference}) to $1$. While anchor merging, the semantic affinity uses scale $\sigma_{\mathrm{sa}}=0.07$, and the spatial affinity uses $\sigma^2_{\mathrm{sp}}=0.04$ with spatial weight $\lambda_{\mathrm{sp}}=1.0$ and residual scalar $\beta = 4.0$.

For LLaVA-NeXT, we apply the official any-resolution preprocessing to all methods and subsequently encode each localized crop as a single \(336 \times 336\) view, yielding $576$ candidate tokens. For Qwen3-VL, we preserve the input aspect ratio while constraining the image area to approximately \(1,152^2\) pixels. Grounding DINO~\citep{liu2024grounding} is the default localizer, with only its grounding heads updated. 
The default rate coefficient is $\lambda_{\mathrm{rate}}=0.02$ for LLaVA-NeXT and $2\times10^{-5}$ for Qwen3-VL. We optimize RUTA using AdamW~\citep{loshchilov2019decoupled} with a learning rate of \(1\times10^{-5}\), a weight decay of \(1\times10^{-4}\), and a batch size of \(6\) per GPU across \(8\) GPUs. Training runs for $2$ epochs on VQAv2 and GQA and $4$ epochs on the remaining datasets.

\subsection{Main Results}

\begin{figure}[!t]
	\centering
\includegraphics[width=.8\textwidth]{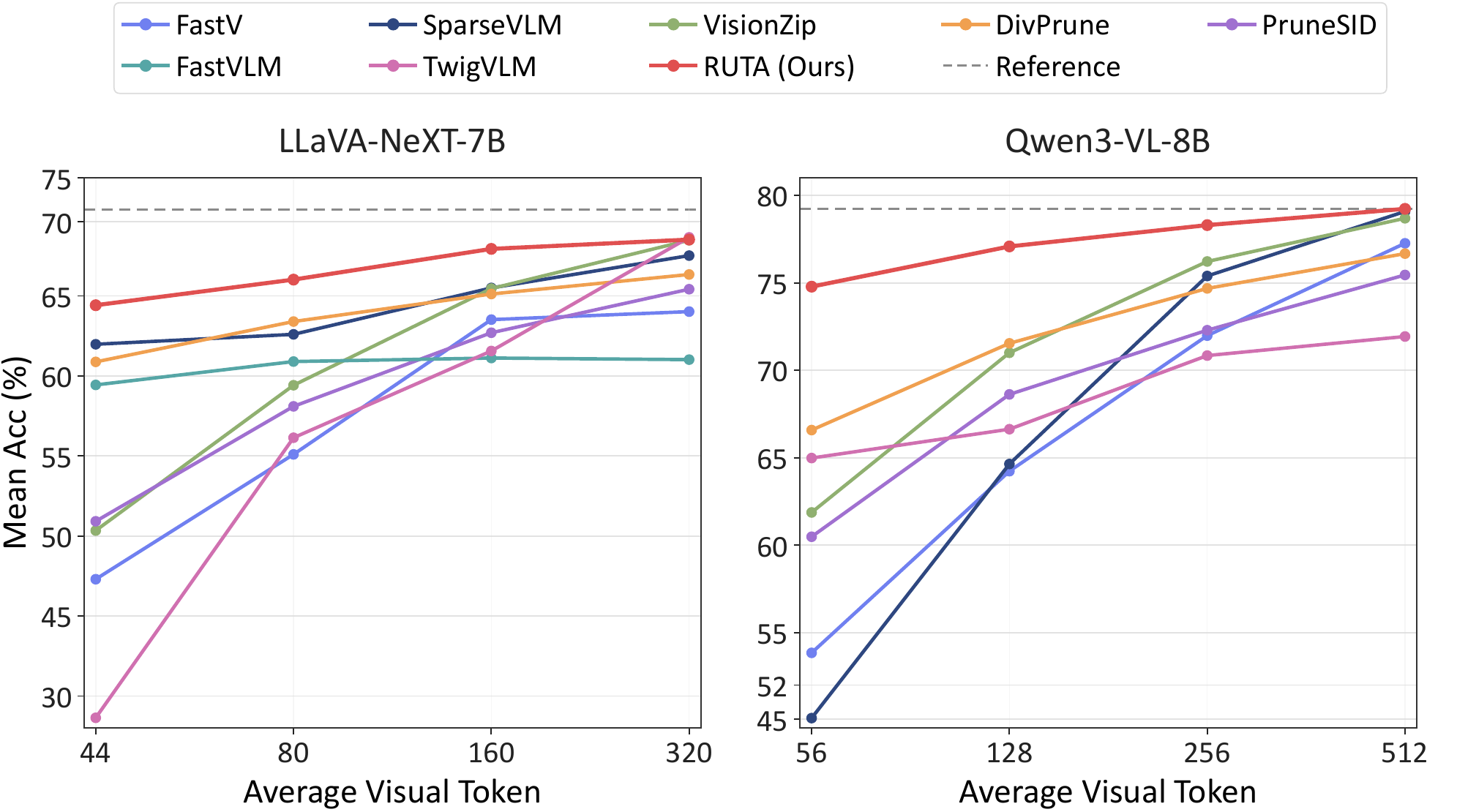}
	\caption{
    Rate–utility curves for LLaVA-NeXT-7B and Qwen3-VL-8B. Each point reports the mean accuracy across the five benchmarks against the corresponding average number of visual tokens processed per sample and LLM layer. Dashed lines indicate the performance of the unreduced reference models. RUTA achieves its largest gains under stringent token budgets, although its advantage narrows on LLaVA-NeXT as the budget increases.
    }
	\label{fig:main_results_all_budget}
\end{figure}

We compare RUTA with training-free reducers based on query relevance or attention (FastV, \citeauthor{chen2024image}, \citeyear{chen2024image}; SparseVLM, \citeauthor{zhang2025sparsevlm}, \citeyear{zhang2025sparsevlm}), similarity-aware merging (VisionZip, \citeauthor{yang2025visionzip}, \citeyear{yang2025visionzip}), and diversity-aware selection (DivPrune, \citeauthor{alvar2025divprune}, \citeyear{alvar2025divprune}; PruneSID, \citeauthor{fang2026prune}, \citeyear{fang2026prune}).
We also include the training-based methods (FastVLM, \citeauthor{vasu2025fastvlm}, \citeyear{vasu2025fastvlm}; TwigVLM, \citeauthor{shao2025growing}, \citeyear{shao2025growing}).

Table~\ref{tab:main_results} supports two main observations. First, RUTA ranks first in mean accuracy for both VLMs and remains close to the best reduced method on every benchmark, indicating stable low-rate performance across different token layouts and evidence demands. Second, the competing methods reveal a mismatch between common reduction criteria and query-specific evidence needs. FastV and SparseVLM derive importance from LLM attention. SparseVLM's contrasting behavior across the two backbones suggests that such signals can be sensitive to token layout and attention patterns. VisionZip, DivPrune, and PruneSID improve preservation through semantic similarity, diversity, and image-complexity cues, respectively, but these largely image-centric criteria do not guarantee query relevance: visually distinctive or complex content may contribute little to the answer. FastVLM and TwigVLM instead target broader system efficiency, so their designs are not optimized specifically for task utility at a matched token rate. After all, reducing spatial resolution can discard fine-grained evidence, while the effectiveness of early-layer pruning may vary across VLM architectures. RUTA addresses these limitations by jointly learning where to focus, which tokens to retain, how many tokens to allocate to each image-query pair, and how to aggregate the remainder under a rate-utility objective. The stronger and more consistent results are therefore aligned with matching both the visual content and the quantity of retained evidence to the current query.

Fig.~\ref{fig:main_results_all_budget} broadens the comparison beyond one low-rate operating point. RUTA leads by a clear margin under aggressive compression on both backbones. On LLaVA-NeXT, however, the curves converge as the budget grows, and RUTA ranks second at $320$ tokens behind PruneSID. This pattern suggests that localized candidate construction is most beneficial when token capacity is scarce. With larger budgets, full-image reducers can preserve broader spatial coverage. On Qwen3-VL, RUTA remains above the competing reducers from $56$ to $512$ tokens and approximately matches the unreduced backbone at the largest setting.

\begin{figure}[!t]
	\centering
    \includegraphics[width=.9\textwidth]{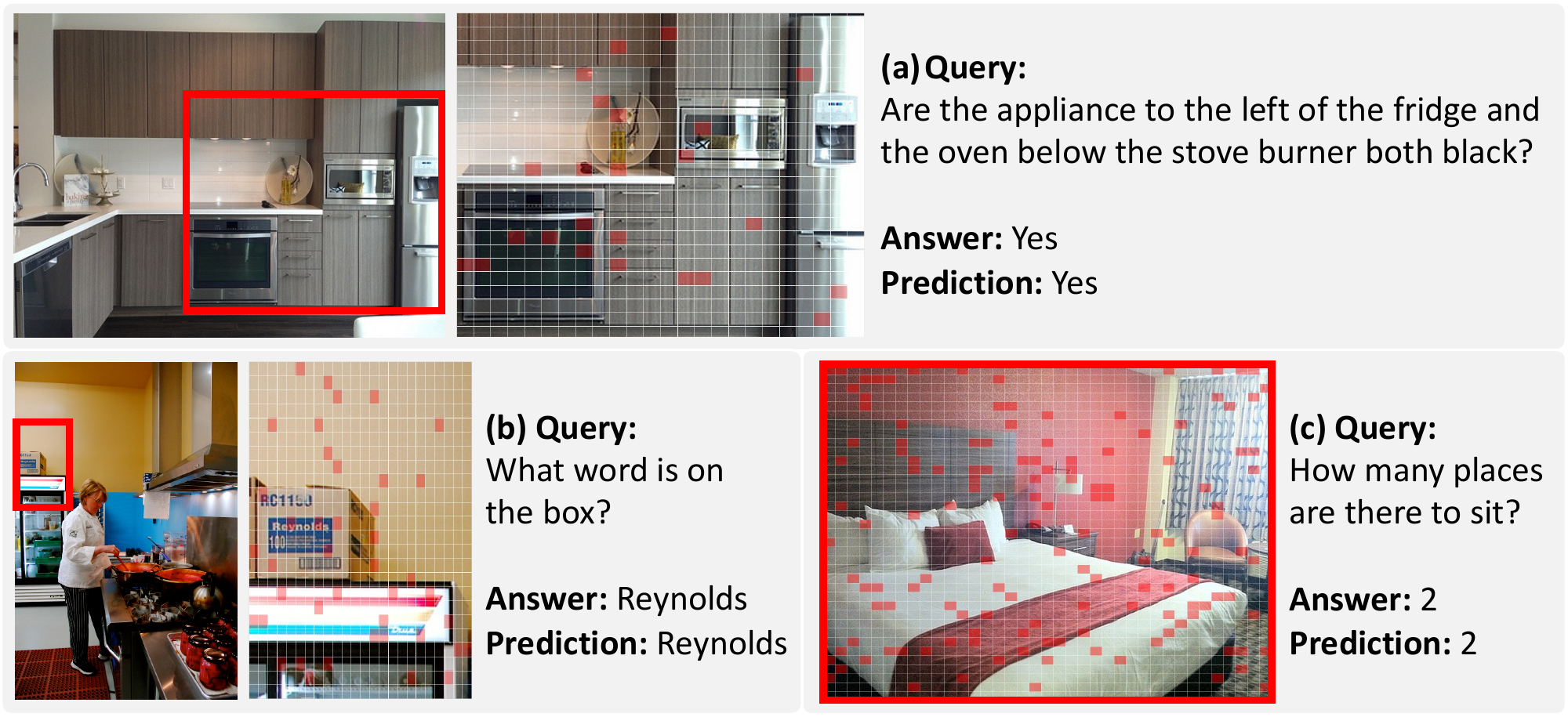}
	\caption{
    Qualitative examples of query-conditioned localization and adaptive token allocation. Red boxes indicate the regions selected by the localizer, and red grid cells denote retained anchors. In \textbf{(a)}, RUTA converts $2,340$ visual tokens from the unreduced LLaVA-NeXT input into $576$ localized candidates and retains $24$ anchors. In \textbf{(b)}, it similarly converts $2,160$  LLaVA-NeXT input tokens into $576$ candidates and retains $42$ anchors. In \textbf{(c)}, the localizer returns no region, so the full image is encoded into $1,344$ candidates, of which $139$ are retained. These examples illustrate how RUTA adapts both the region examined and the number of retained tokens to the query.
    }
	\label{fig:visualization}
\end{figure}

\paragraph{Visualizations} 
Fig.~\ref{fig:visualization} illustrates how candidate construction and rate allocation respond to different evidence demands. In (a), the query requires several appliances and their spatial relations; the localized region then covers the oven, refrigerator, and stove burner while excluding nearly half of the image. In (b), localization magnifies the text-bearing box before the compressor selects a small anchor set. In (c), the query provides no explicit region to ground, so RUTA uses the full image and allocates more anchors to cover the scene, which extend beyond the immediate answer locations. This spatial spread supports the merge in Eq.~(\ref{eq:anchor_residual_merge}), allowing each anchor to absorb information from nearby non-anchors while preserving contextual coverage.

\subsection{Ablation Studies}

\begin{figure}[!t]
	\centering
    \includegraphics[width=.9\textwidth]{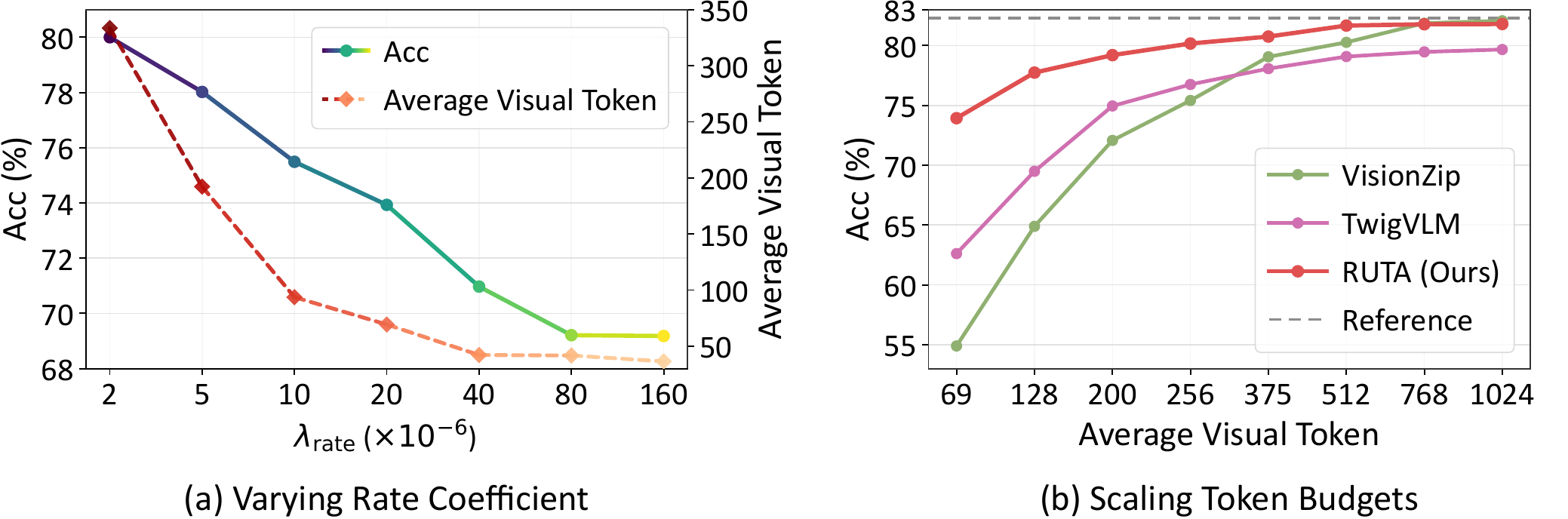}
    \caption{Rate control and token budget scaling on TextVQA with Qwen3-VL-8B. Visual token counts are averaged over samples and LLM layers. \textbf{(a)} Increasing the rate coefficient $\lambda_{\mathrm{rate}}$ progressively reduces token usage, illustrating the resulting rate-utility trade-off. \textbf{(b)} Accuracy as a function of the average visual token count for VisionZip, TwigVLM, and the proposed RUTA. The dashed line marks the unreduced reference model. RUTA performs particularly well in the low-budget regime and approaches the reference accuracy as the budget increases.}
\label{fig:ablation_lambda_scaling}
\end{figure}

\begin{figure}[!t]
	\centering
    \includegraphics[width=.9\textwidth]{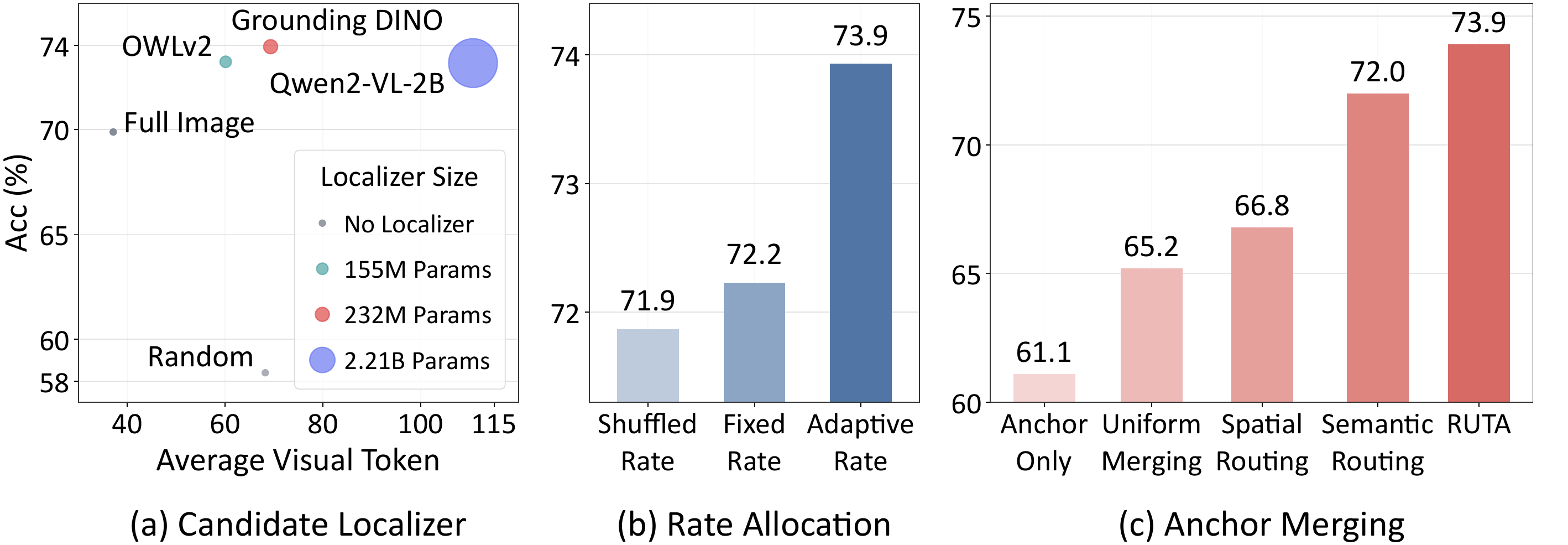}
    \caption{Component ablations on TextVQA with Qwen3-VL-8B. \textbf{(a)} Comparison of query-conditioned candidate localizers, along with random-region and full-image degenerates. Marker area is proportional to the localizer parameter count. \textbf{(b)} Comparison of fixed, shuffled, and adaptive rate allocation at approximately $70$ visual tokens on average. \textbf{(c)} Comparison of anchor-only retention and four merging strategies at a matched average rate. 
    }
    \label{fig:ablation_localizer_allocation_merging}
\end{figure}

We conduct five controlled studies on TextVQA with Qwen3-VL-8B. We first verify that the rate coefficient controls the operating point, then examine how the advantage changes with the token budget, and finally isolate the contributions of candidate construction, adaptive token allocation, and anchor merging. 

\paragraph{Effect of the rate coefficient}
Fig.~\ref{fig:ablation_lambda_scaling}(a) shows that $\lambda_{\mathrm{rate}}$ directly controls the rate-utility operating point. Increasing it from $2\times10^{-6}$ to $2\times10^{-5}$ reduces the average rate from $333.6$ to $69.3$ tokens, while accuracy decreases from $80.0\%$ to $73.9\%$. Beyond this point, stronger rate pressure produces progressively smaller rate reductions while accuracy falls toward $69.2\%$. We therefore use $\lambda_{\mathrm{rate}}=2\times10^{-5}$ for Qwen3-VL because it lies near the knee of the curve, providing a substantial rate reduction before utility declines more sharply.

\paragraph{Scaling token budgets}
Fig.~\ref{fig:ablation_lambda_scaling}(b) compares VisionZip, TwigVLM, and RUTA across expanded token budgets. RUTA remains above TwigVLM throughout the evaluated range. The narrowing margin indicates that query-conditioned candidate construction is most valuable when representational capacity is scarce. VisionZip becomes comparable at $768$ tokens and slightly surpasses RUTA at $1{,}024$ tokens, where a full-image reducer can preserve broader context directly. Thus, RUTA’s main advantage lies in low-to-medium-rate operation, which is particularly relevant to practical deployment because meaningful reductions in LLM-side computation and memory require operating well below the unreduced token count.

\paragraph{Choice of candidate localizer}
Fig.~\ref{fig:ablation_localizer_allocation_merging}(a) varies the candidate localizer while keeping the compressor and training protocol fixed. OWLv2~\citep{minderer2023scaling}, Qwen2-VL-2B~\citep{wang2024qwen2}, and Grounding DINO~\citep{liu2024grounding} achieve comparable accuracy and consistently outperform the random-region and full-image controls. This consistency across architectures and parameter scales suggests that the gains arise primarily from query-conditioned localization rather than a specific localizer. Because the variants produce different average token rates, the results demonstrate robustness to localizer choice rather than a strictly rate-matched ranking. We use Grounding DINO by default because it achieves the highest accuracy with a moderate parameter count.

\paragraph{Effect of adaptive rate allocation}
Fig.~\ref{fig:ablation_localizer_allocation_merging}(b) separates rate variation from demand-aware assignment. The fixed-rate control gives every sample $70$ tokens, whereas the shuffled-rate control preserves RUTA's rate distribution but randomly reassigns those rates across samples. At the same average rate, adaptive allocation reaches the highest accuracy. The gain over fixed rate shows that varying the budget across samples is useful, while the gain over shuffled rate shows that the budget must also be assigned to the appropriate input. Rate variation alone is therefore insufficient; its value comes from matching capacity to evidence demand.

\paragraph{Effect of anchor merging} 
Finally, Fig.~\ref{fig:ablation_localizer_allocation_merging}(c) fixes the selected anchors and average rate to isolate how non-anchors should be incorporated. Uniform merging, where $w_{i,j}$ in Eq.~(\ref{eq:anchor_merge_assignment_2}) is set to $1/|\mathcal{A}|$, improves over retaining anchors alone, showing that non-anchors contain complementary evidence. Semantic routing provides a substantially larger benefit than spatial routing, indicating that content affinity is the primary assignment signal. Their combination performs best, confirming that spatial proximity provides complementary structure when routing non-anchor information to semantically compatible anchors. These results motivate the joint semantic-spatial formulation in Eq.~(\ref{eq:anchor_merge_assignment_1}).

\section{Conclusion and Discussion}

We have presented RUTA, a rate-utility optimization approach that learns both which visual evidence to retain and how many tokens each image-query pair should pass to the LLM. RUTA combines query-conditioned candidate construction, probabilistic token retention, and anchor-based aggregation, while its rate penalty provides direct control over the accuracy-efficiency trade-off. Across five VQA benchmarks and two VLM architectures, RUTA achieves the highest mean accuracy among the reduced methods at matched stringent token budgets. These results suggest that adapting both the  visual content and the quantity of retained evidence is particularly valuable when token capacity is scarce.

The current formulation has three practical limitations. First, visual token count is a model-relative proxy for efficiency and does not capture localization and encoding overhead, end-to-end latency, KV-cache memory, or hardware-dependent throughput. Second, candidate construction uses a single enclosing region to preserve intervening context and spatial relationships among grounded regions. When these regions are widely separated, however, the resulting crop may cover much of the image and thereby reduce the resolution benefit of localization. Performance may also degrade when relevant evidence is small or not reliably
identified by the localizer. Third, RUTA is optimized for a particular VLM and task dataset; deployment on a new backbone or domain therefore requires additional training, reducing its plug-and-play appeal. Our evaluation on single-image VQA and two VLM architectures also leaves broader transfer to be established.

A particularly valuable next step is a universal reducer that can be trained once and attached to different frozen VLMs with minimal or no task-specific fine-tuning. Backbone-agnostic token interfaces, self-supervised or distillation-based objectives, and lightweight calibration adapters could make learned adaptive allocation as convenient to deploy as training-free pruning. A complementary direction is to condition a single reducer on an inference-time token, latency, or memory target, allowing it to span multiple rate-utility operating points and device profiles without retraining for each setting. More broadly, adaptive allocation could become a resource-scheduling mechanism across regions, images, and video frames, facilitating multimodal systems to decide not only what evidence to preserve but also where limited computation should be spent over time.


\bibliographystyle{assets/plainnat}
\bibliography{ref}

\end{document}